\documentclass[11pt]{article}
\usepackage{coling2020}
\usepackage{times}
\usepackage{url}
\usepackage{multirow}
\usepackage{hyperref}
\usepackage{graphicx}
\usepackage{amssymb}

\colingfinalcopy 

\title{Token Drop mechanism for Neural Machine Translation}

\author{Huaao Zhang \\
  Soochow University / Suzhou China \\
  {\tt hazhang@stu.suda.edu.cn} \\
  Xiangyu Duan\thanks{* Corresponding author.}\\
  Soochow University / Suzhou China \\
  {\tt xiangyuduan@suda.edu.cn} \\\And
  
  Shigui Qiu \\
  Soochow University / Suzhou China \\
  {\tt sgqiu@stu.suda.edu.cn} \\
  Min Zhang \\
  Soochow University / Suzhou China \\
  {\tt minzhang@suda.edu.cn} \\
  }

\date{}

\begin{document}
\maketitle
\begin{abstract}

  Neural machine translation with millions of parameters is vulnerable to unfamiliar inputs. We propose Token Drop to improve generalization and avoid overfitting for the NMT model.  Similar to word dropout, whereas we replace dropped token with a special token instead of setting zero to words.  We further introduce two self-supervised objectives: Replaced Token Detection and Dropped Token Prediction. Our method aims to force model generating target translation with less information, in this way the model can learn textual representation better. Experiments on Chinese-English and English-Romanian benchmark demonstrate the effectiveness of our approach and our model achieves significant improvements over a strong Transformer baseline\footnote{Our code is released at \url{https://github.com/zhajiahe/Token_Drop}}.

\end{abstract}

\section{Introduction}
\label{intro}

%
%
\blfootnote{
    %
    %
    %
    %
    %
    This work is licensed under a Creative Commons 
    Attribution 4.0 International License.
    License details:
    \url{http://creativecommons.org/licenses/by/4.0/}.
}
Neural machine translation (NMT) achieved enormous success in advancing the quality of translation \cite{Bahdanau2015Neural,Vaswani2017,pmlr-v70-gehring17a}. In spite of the impressive performance, NMT models are still vulnerable to perturbations in the input sentences \cite{belinkov2018synthetic,cheng-etal-2019-robust} , i.e. a tiny perturbation will affect hidden representation and lead to low quality of translation \cite{zhao2017generating}. Moreover, NMT commonly consists of  millions of parameters, which making it prone to overfitting especially in low resource scene.

A natural way to improve generalization is synthesizing natural noise \cite{karpukhin-etal-2019-training} or adopting arbitrary noise \cite{cheng-etal-2018-towards,ebrahimi-etal-2018-adversarial}.  Another way is exploring regularization techniques to avoid overfitting \cite{miceli-barone-etal-2017-regularization}, making model robust to unseen or unfamiliar inputs. However, as discrete data, the text is hard to retain the semantic information after corruption.

In this paper, we propose Token Drop to prevent overfitting and improve generalization. Different from standard dropout \cite{srivastava2014dropout} that drops neurons in network randomly, we drop tokens of the input sentences. In order to retain semantic information, we replace tokens with a special symbol $<\textit{unk}>$ . This allows model learn hidden representation from rest token's context, and predict target translation condition on latent variable. On the one hand, our method allows model meeting exponentially different sentences can be explained as data augmentation; On the other hand, our method corrupts input sentences with natural noise can be seen as regularization term for NMT.
 
We investigate two self-supervised objectives: Replaced Token Detection and Dropped Token Prediction. Considering  our Token Drop method regularize parameters by weakening model inputs, making NMT suitable for applying self-supervised objective. During training: (1) use a discriminator to detect whether input tokens are dropped or not; (2) leverage hidden state to predict original tokens of dropped tokens inspired by \textit{Cloze} task \cite{devlin-etal-2019-bert}. Both of them guide model to generate semantically similar representation, leading to a better generalization capacity. 
\section{Token Drop Training}
Standard dropout prevents overfitting by setting input neurons or hidden neurons to zero with a certain probability $p$ \cite{hinton2012improving,srivastava2014dropout}. whereas we consider the input sequences of machine translation models instead of the network's neurons, which named Token Drop. Given a input tokens sequence of sentence $X = \{ x_{1},x_{2},...,x_{n} \} $ and posit a $|X|$ independent drop rate $p$. The token $x_{i}$ in $X$ will be dropped if $m_{i}$ is 1 . This process as Equation~\ref{bernoulli}:
\begin{equation}\label{bernoulli}
    \textbf{m} \sim Bernoulli(p) \qquad \hat{X}= \textbf{REPLACE}(X, \textbf{m}, <\textit{unk}>)
\end{equation}
\begin{equation}
    \mathcal{L}_M = -\log \sum P(Y|\hat{X},\theta_{M})
\end{equation}
The Token Drop can be interpreted as data augmentation and regularization technique for NMT. Seeing that NMT model commonly adopts encoder and decoder architecture, therefore our method drops tokens for both source and target inputs. For the source side, model encoder learns intermediate representation by exponentially different incomplete sentences. For the target side, model decoder generates target translation condition on latent variable, weakening the constraint caused by teacher forcing. Both of them receives incomplete information from inputs, simulating the real situation (e.g unknown or unfamiliar data) at test time. 
\subsection{Token Drop methods}
We adopt three drop strategy for Token Drop:





\textbf{Zero-Out}
is introduced by  \newcite{sennrich-etal-2016-edinburgh}, different from the standard dropout, the method drops full word by setting zero to word embedding during training. The deficiency is  zero vector can not learn representation from its context in the self-attention layer.

\textbf{Drop-Tag}
 \cite{kageback-salomonsson-2016-word} replaces token with a {$<\textit{dropped}>$} tag. The tag is subsequently treated just like any other word in the vocabulary and has a corresponding word embedding that is trained. We adopt this technique for NMT to learn better feature representation. 
 
\textbf{Unk-Tag} replaces token with generic unknown word token 
    {$<\textit{unk}>$}.  \newcite{bowman-etal-2016-generating} and \newcite{pmlr-v70-yang17d} apply it to RNN decoder to force model make prediction by latent variable. We found this perfectly suits for NMT system especially on self-attention layers. Better than Drop-Tag method, it need not to add an extra token as well as parameters.

\subsection{Replaced Token Detection}
We propose the Replaced Token Detection task to promote generalization ability of the model encoder. We regard dropped information as self-supervised label, following  \newcite{Clark2020ELECTRA:}, we train a discriminator $D(G(x))$ to detect whether tokens are dropped or not. On account of our dropped tokens are obvious to distinguish, so we add a simple linear classifier to detect replaced tokens. The objective is :
\begin{equation}
    \mathcal{L}_{RTD} = \mathbb{E}_{x \sim \mathcal{S}}[-\log D(G(\mathbf{x}))] + \mathbb{E}_{\hat{x} \sim d(x)}[1 - \log D(G(\hat{x}))]
\end{equation}
Where $d(x)$ denotes the dropped tokens. In our model, the encoder serves as a generator $G$, which generates hidden state of input tokens. The discriminator $D$ tries to distinguish whether a token is dropped or not, while the generator $G$ has to produce similar representation for $x$ and $\hat{x}$, making the model robust to noisy and unknown inputs.

\subsection{Dropped Token Prediction}
In consideration of  our Token Drop model randomly replaces tokens of input sentence, similar to Masked Language Model \cite{devlin-etal-2019-bert}, which masks then predicts masked tokens by the rest of the tokens, making use of contextual information. Accordingly, we propose Dropped Token Prediction (DTP), predicting dropped tokens by their hidden states. The DTP objective is : 
\begin{equation}
    \mathcal{L}_{DTP} =   -  \log \sum_{\hat{x} \in d(x)} P(\hat{x}|X_{/d(x)}, \theta_M)
\end{equation}
\begin{equation}
    P(\hat{x}|X_{/d(x)}) = E(G(X_{/d(x)}))
\end{equation}
Where $d(x)$ and $X_{/d(x)}$ denote the dropped tokens and the rest tokens respectively. $G$ is model encoder, $E(.)$ is prediction layer.  In our implementation of DTP, we adopt \textit{weight tying} \cite{press-wolf-2017-using} , that is to share the same weight matrix between embedding layer and token prediction classifier. 

At the end we train our model jointly with DTP and RTD objective:
\begin{equation}
    \mathcal{L} = \mathcal{L}_M + \alpha \mathcal{L}_{RTD} + \beta \mathcal{L}_{DTP}
\end{equation}

\section{Experiment}
We conduct our approach on two machine translation benchmarks: LDC (ZH-EN) and WMT16 (EN-RO)\footnote{http://www.statmt.org/wmt16/}
\subsection{Dataset and Evaluation}
For ZH-EN translation, we used 1.25M sentence pairs extract from LDC corpus. Byte-pair encoding is employed separate vocabulary of about 42K and 31K tokens with 32K merge operations \cite{sennrich-etal-2016-neural}. we chose NIST06 as the valid set and NIST02, NIST03, NIST04, NIST05, NIST08 as the test set, which contains 878, 919, 1788, 1082, 1357 sentence pairs respectively. We measure BLEU score with \textit{multi-bleu.pl}\footnote{https://github.com/moses-smt/mosesdecoder/blob/master/scripts/generic/multi-bleu.perl}. For WMT16 EN-RO data which consists of 610K pairs, we adopt  \newcite{lee-etal-2018-deterministic}'s preprocessing. The vocabulary is 35K joint source and target subwords \cite{sennrich-etal-2016-neural}.  we use newstest-2016 as test set and  report tokenized BLEU score.
\subsection{Models and Settings}
We    adopt    the    Transformer    model \cite{Vaswani2017} implemented  in  PyTorch  in  the \textit{fairseq-py} toolkit \cite{ott-etal-2019-fairseq}. We closely followed settings by \newcite{Vaswani2017}$(d_{model} = 512, d_{hidden} = 512, d_{FFN}= 2048, n_{layer} = 6, n_{head} = 8)$ and used dropout of $p_{dropout} = 0.3$.   As for our approach, we set drop rate $p_s=0.15$ and $p_t=0.3$ respectively for ZH-EN task. For EN-RO, we set $p_s=0.15$ and $p_t=0.2$. To train with DTP and RTD objective, we simply set $\alpha,\beta = 1 $ .
\section{Result}
\begin{table}[h]
\begin{center}
    \centering
    \begin{tabular}{l||cccccc||cc}
    \hline
    \multirow{2}{*}{Models} & \multicolumn{6}{|c||}{LDC ZH-EN} & \multicolumn{2}{c}{WMT EN-RO} \\
    \cline{2-9} 
     & NIST02 & NIST03 & NIST04 & NIST05& NIST08 & AVG & EN$\rightarrow $RO & RO$\rightarrow$ EN\\
    \hline
    Transformer & 47.17 & 46.78 & 47.46 & 47.97 & 38.01 & 45.47 & 34.60 & 33.96 \\ 
     Zero-Out & 47.86 & 47.12 & 48.86 & 47.90 & 38.65 & 46.08 & 35.21 & 35.00\\
     Drop-Tag & 48.99 & 48.50 & 49.81 & 49.60 & 39.31 & 47.24 & 35.34 & 35.19\\
     Unk-Tag & 48.96 & 48.52 & 49.50 & 49.49 & 39.31 & 47.16 & 35.28 & 35.13\\
    + RTD & 49.00 & \textbf{49.36} & 49.43 & 49.25 & 39.92 & 47.39 & 35.49 & 35.38\\
    + DTP & 49.16 & 49.19 & 49.99 & \textbf{49.83} & 40.20 & 47.75 & \textbf{35.75} & 35.38\\
    + RTD\&DTP & \textbf{49.52} & 49.29 & \textbf{50.17} & 49.82 & \textbf{40.38} & \textbf{47.84} & 35.67 & \textbf{35.69} \\
    \hline
    \end{tabular}
    \label{tab:ldc_result}
    \caption{\label{Result} BLEU scores on test set for LDC Chinese-English and WMT16 English-Romanian tasks}
\end{center}
\end{table}
The results of our experiment on NIST Chinese-English and WMT16 English-Romanian tasks are shown in Table~\ref{Result}. We first conduct Token Drop through three drop methods (Zero-out, Drop-Tag, Unk-Tag), the results show that Token Drop model significantly outperform baseline on two languages. Furthermore, we combine Unk Tag method with DTP and RTD training objective, the results show that both DTP and RTD provide a further improvement on Token Drop training. Overall, we get a gain of 2.37, 1.15 and 1.73 BLEU score on three tasks respectively.
\begin{table}[h]
\begin{center}
    \centering
    \begin{tabular}{c||c|c|c|c}
    \hline
    Method & 0.00 & 0.05 & 0.10 & 0.15 \\
    \hline \hline
    Beseline & 47.44 & 39.26 & 30.83 & 23.20 \\
    \cite{cheng-etal-2019-robust} & 46.95 & 44.20 & 41.71 & 39.89 \\
    \hline 
    Ours & \textbf{48.75} & \textbf{46.55} & \textbf{44.12} & \textbf{41.64} \\
    \hline
    \end{tabular}
    \caption{\label{noise} Result on incomplete inputs with different $p_s$ for ZH-EN valid set (NIST06).}
\end{center}
\end{table}

In order to demonstrate the generalization capacity of our model on real situation. We constrain input information by replacing words with generic unknown symbol $<\textit{unk}>$.  For each sentence, we generate 100 noisy sentences then report average BLEU score. We also compare our method with \newcite{cheng-etal-2019-robust}, who introduced white-box adversarial noisy inputs to improve robustness. Table~\ref{noise} reports our result on incomplete inputs, from where we can see our approach outperforms previous method. This improvement confirms that our model obtains larger generalization capacity over baseline. 
\begin{figure}[h]
\begin{minipage}[b]{.5\linewidth}
\centering
\includegraphics[width=8cm]{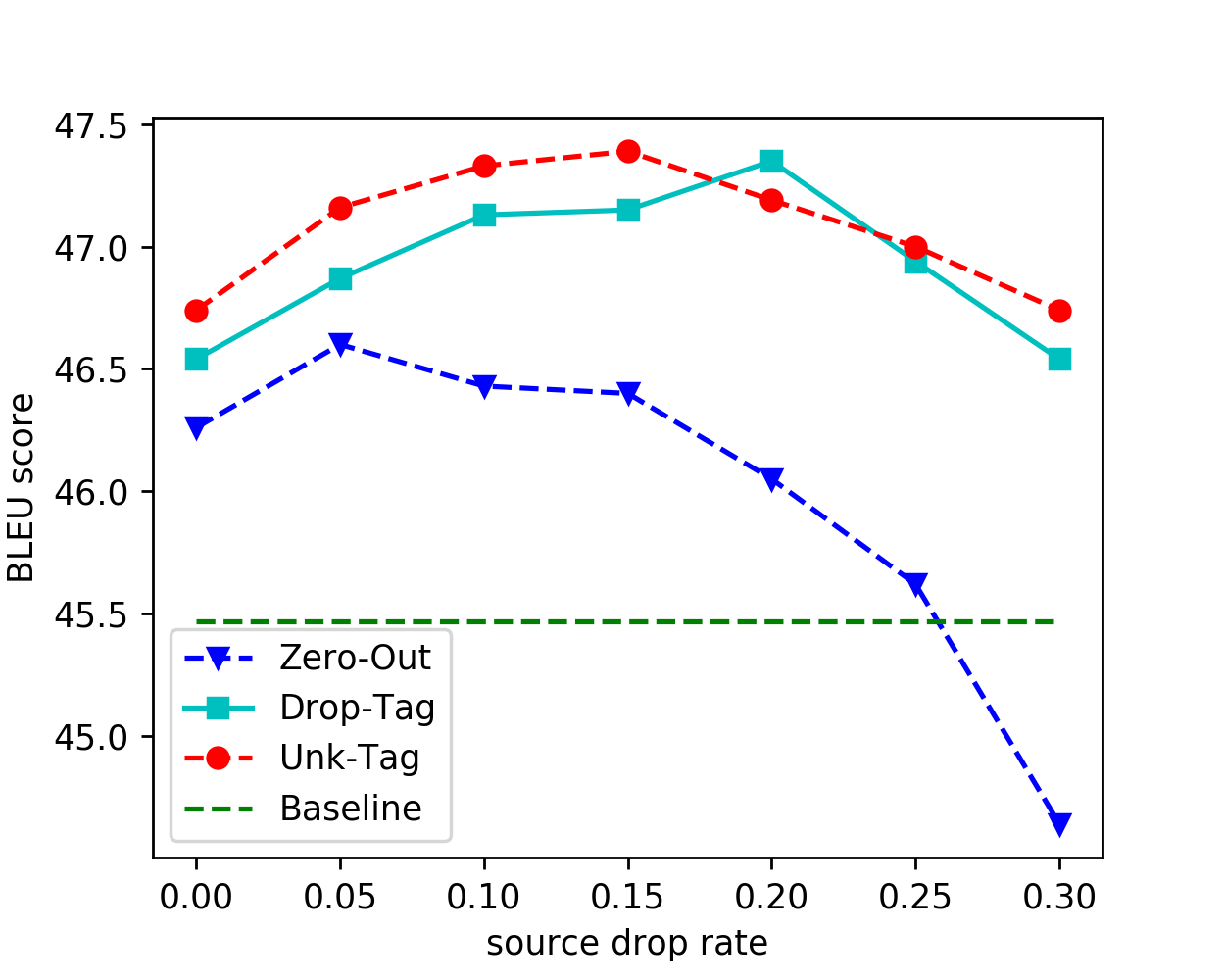}
\caption{\label{Drop_rate} Effect of different source drop rate  $p_s$\\ ($p_t = 0.3$) on LDC ZH-EN translation task}
\end{minipage}%
\begin{minipage}[b]{.5\linewidth}
\centering
\includegraphics[width=8cm]{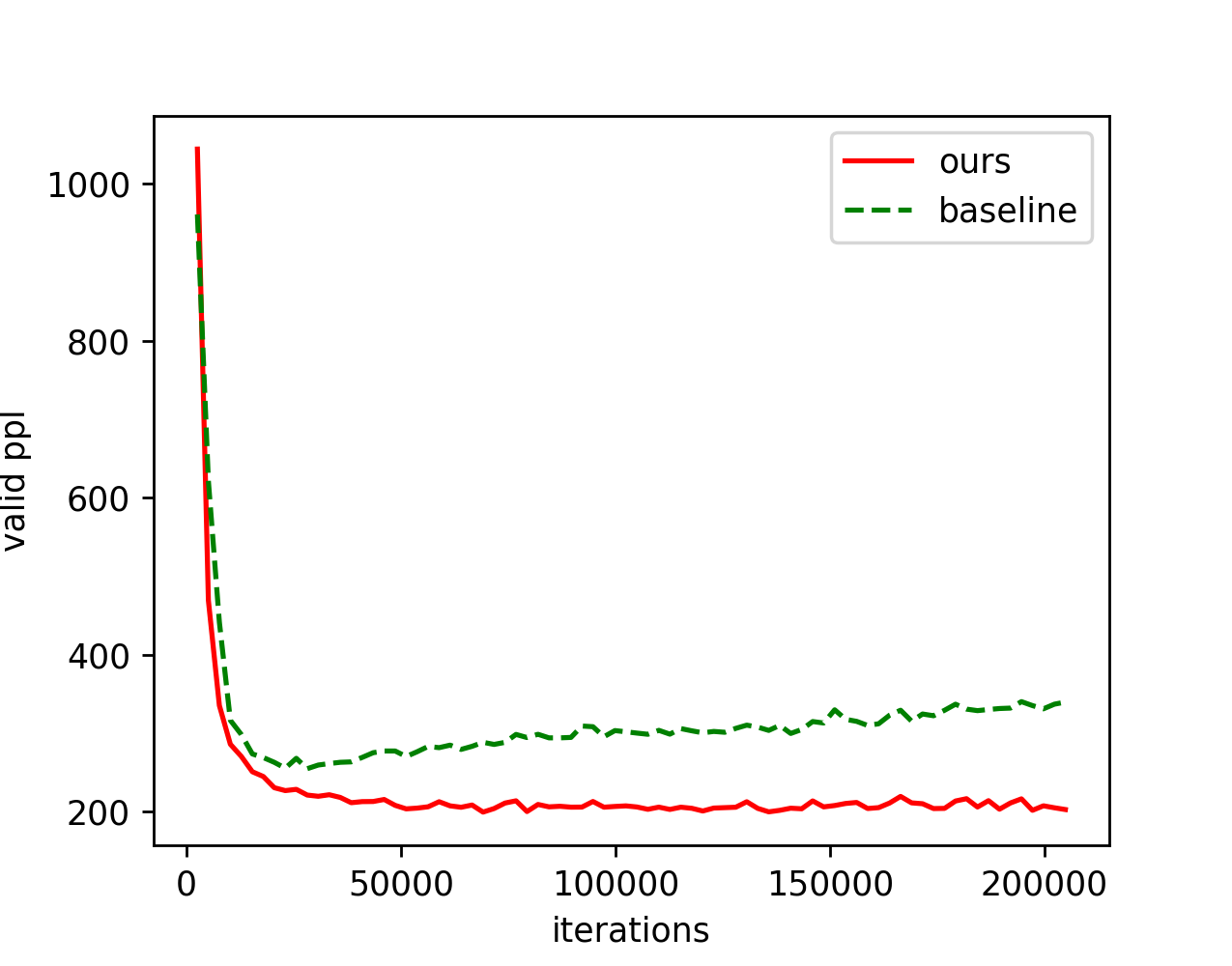}
\caption{\label{Learning_curve} Learning curves of baseline model \\ of our Token Drop model}
\end{minipage}%
\end{figure}

To examine the impact of Token Drop, we train our model with different source drop rate $p_s=[0.0,0.05,0.1,0.15,0.2,0.25, 0.3]$. From Figure~\ref{Drop_rate} we can see that model training with a moderate drop rate $p$ would advanced in performance significantly. Drop-Tag and Unk-Tag are quite similar, both of them outperforms Zero-Out method. We plot learning curves of baseline model and our Token Drop model. Figure~\ref{Learning_curve} shows that with the increase of training iterations, our model achieves lower and more stable perplexity than baseline, demonstrating the effectiveness of our approach to prevent overfitting and improve translation quality.

\section{Related Work}
 

\textbf{Word Dropout}\quad \newcite{iyyer-etal-2015-deep} proposed \textit{word dropout} as feature extractor for text classification task. \newcite{bowman-etal-2016-generating} and \newcite{xie2017data} applied \textit{word dropout} to RNN decoder can be regard as a smoothing technique. For machine translation task, \newcite{sennrich-etal-2016-edinburgh} randomly set zero to words of input sentence to prevent overfitting, advancing in considerable performance on noisy dataset. In this paper, we explain word dropout as data augmentation (i.e. allows model meeting exponentially different sentences) and a regularization technique (i.e. weakens the encoder and decoder, obtaining better intermediate representations).

\section{Conclusion}
In this paper, we have proposed Token Drop mechanism for neural machine translation task. Inspired by self-supervised learning, we introduced Replaced Token Detection and Dropped Token Prediction training objective. We found that NMT model trained with Token Drop gains larger generalization capacity and reduction in overfitting. Even without prior knowledge and additional parameters, our proposed approach reports convincing results on neural machine translation. In future work, we plan to investigate impact of dropping on different words, e.g. word importance and word type. 

\section*{Acknowledgements}
The authors would like to thank the anonymous reviewers for the helpful comments. This work was supported by National Natural Science Foundation of China (Grant No. 61673289, 61525205).

\bibliographystyle{coling}
\bibliography{coling2020}
\end{document}